\documentclass[12pt, a4paper]{article}

\usepackage[T1]{fontenc}
\usepackage[utf8]{inputenc}
\usepackage{lmodern}
\usepackage{microtype}
\usepackage{amsmath, amssymb, amsthm}
\usepackage{asymptote}
\usepackage{geometry}
\usepackage{hyperref}
\usepackage{xcolor}
\usepackage{mdframed}
\usepackage{enumitem}
\usepackage{titlesec}
\usepackage{titling}
\usepackage{abstract}
\usepackage{tocloft}
\usepackage{booktabs}
\usepackage{graphicx}
\usepackage{float}
\usepackage{longtable}
\usepackage{dsfont}

\definecolor{darkblue}{RGB}{25, 50, 120}
\definecolor{blockbg}{RGB}{245, 247, 252}
\definecolor{blockrule}{RGB}{100, 130, 200}
\definecolor{notebg}{RGB}{250, 248, 235}
\definecolor{noterule}{RGB}{190, 160, 60}

\hypersetup{
  colorlinks=true,
  linkcolor=darkblue,
  urlcolor=darkblue,
  citecolor=darkblue,
  pdftitle={Trust via Reputation of Conviction},
  pdfauthor={Aravind R. Iyengar},
}

\titleformat{\section}
  {\large\bfseries\color{darkblue}}{\thesection.}{0.6em}{}[\vspace{-0.3em}\rule{\linewidth}{0.4pt}]
\titleformat{\subsection}
  {\normalsize\bfseries\color{darkblue}}{\thesubsection.}{0.5em}{}
\titleformat{\subsubsection}
  {\normalsize\itshape\bfseries}{\thesubsubsection.}{0.5em}{}

\titlespacing{\section}{0pt}{1.6em}{0.6em}
\titlespacing{\subsection}{0pt}{1.2em}{0.4em}
\titlespacing{\subsubsection}{0pt}{1.0em}{0.3em}

\newmdenv[
  backgroundcolor=blockbg,
  linecolor=blockrule,
  linewidth=2pt,
  topline=false, rightline=false, bottomline=false,
  innerleftmargin=10pt, innerrightmargin=8pt,
  innertopmargin=6pt, innerbottommargin=6pt,
  skipabove=8pt, skipbelow=8pt,
]{remark}

\newmdenv[
  backgroundcolor=notebg,
  linecolor=noterule,
  linewidth=2pt,
  topline=false, rightline=false, bottomline=false,
  innerleftmargin=10pt, innerrightmargin=8pt,
  innertopmargin=6pt, innerbottommargin=6pt,
  skipabove=8pt, skipbelow=8pt,
]{note}

\title{%
  {\LARGE\bfseries\color{darkblue} Trust via Reputation of Conviction}
}
\author{Aravind R.\ Iyengar}
\date{March 8, 2026}

\begin{document}

\maketitle
\thispagestyle{empty}

\vspace{1.5em}
\begin{abstract}
\noindent
The question of \emph{knowledge}, \emph{truth} and \emph{trust} is explored via a mathematical formulation of claims and sources. We define truth as the reproducibly perceived subset of knowledge, formalize sources as having both generative and discriminative roles, and develop a framework for reputation grounded in the \emph{conviction} --- the likelihood that a source's stance is vindicated by independent consensus. We argue that conviction, rather than correctness or faithfulness, is the principled basis for trust: it is regime-independent, rewards genuine contribution, and demands the transparent and self-sufficient perceptions that make external verification possible. We formalize reputation as the expected weighted signed conviction over a realm of claims, characterize its behavior across source-claim regimes, and identify continuous verification as both a theoretical necessity and a practical mechanism through which reputation accrues. The framework is applied to AI agents, which are identified as capable but error-prone sources for whom verifiable conviction and continuously accrued reputation constitute the only robust foundation for trust.
\end{abstract}

\vspace{1em}
\tableofcontents
\vspace{1em}

\section{Knowledge: The Precursor}

We separate the discussion of \emph{knowledge} from \emph{truth} in the following way. Knowledge can be defined as the learning of information. For simplicity, we think of knowledge as being attained via exposure to \emph{claims}. Without knowledge, there is no question of truth. Once one has \emph{known} something, the question of whether it is \emph{true} arises. Once knowledge has been gained, one can typically propagate it via communication; knowledge propagated may also be sometimes different from the knowledge gained.

\section{Truth: The Pursuit}

In the context of what we discussed above, \emph{truth} is the subset of knowledge which is experienced or perceived \emph{objectively}. This notion of truth immediately reveals the following two properties:

\begin{itemize}[leftmargin=1.8em, itemsep=4pt]
  \item \textbf{Perception-dependency}: Truth depends on the means through which it is experienced or perceived. For example, a claim on what an object's surface looks like is truthful or not depending on whether it is seen directly by the eye or under a microscope. This implies that if there are limitations of the method of perception, they have a bearing on what is considered true.

  \item \textbf{Reproducibility-dependency}: The objectivity requirement implies that truth inherently depends on how reproducible or how widely perceptible a claim is. For example, in a hypothetical world where all but one human were blind, the claim that the sun rises from the east may not be held as a truth (ignoring the other ways, e.g., warmth, that the sun can be perceived). Note that reproducibility can either be at different times, e.g., repeated experimentation, or at the same time, e.g., objectively perceptible events.
\end{itemize}

\begin{remark}
\textbf{Truth is a \emph{social} construct.} This definition of truth brings out the property that truth only arises as a concept when there are multiple perceptions involved. A good illustration is that if there were only one human in the world and she were color-blind, that would be the singular perception. Only when there is a second human is there even a question of a \emph{shared} or \emph{objectively perceived} reality, i.e., truth.
\begin{itemize}[leftmargin=1.4em, itemsep=3pt, topsep=4pt]
  \item Science as a pursuit of truth naturally fits into this framework --- we continuously set down claims (hypotheses) and search for reproducible ways to perceive them. Refinements in scientific theory are largely due to improved means of perception becoming accessible over time, both via better physical mechanisms (e.g., microscopes, finer time-measuring devices) and better analytical theories (e.g., relativity, causal inference).
  \item ``Subjective truths'' are not truths, e.g., ``I have a headache'', ``I like orange'' or ``I think we should take this chance''. Such experiences are better classified as \emph{feelings}, \emph{likings} or \emph{opinions}.
  \item ``Imperceptible truths'' are possible, but they are also \emph{unknowable}, e.g., imagine ``the sun rises from the east'' in a hypothetical world where no humans had eyesight.
\end{itemize}
\end{remark}

\subsection{Establishing Truth}
\label{sec:establishing-truth}

When it comes to claims that can be reproduced over time, establishing truth entails being able to conduct repetitive experiments and evaluate the validity of the claims. Such validation is possible via the following mechanisms:

\begin{enumerate}[leftmargin=1.8em, itemsep=4pt]
  \item \textbf{Confirmation}: Establishing truth via repeated experimentation with the same source. Note that such confirmation may be exposed to \emph{systematic errors} or \emph{biases} native to the source of perception.
  \item \textbf{Verification}: Even when a claim is reproducible over time, its perception can be exposed to some degree of subjectivity if experimented by different actors or observed via different sources. Establishing objectivity across sources and repeated experimentation is verification.
\end{enumerate}

For claims that cannot be reproduced over time, the only way to \emph{confirm} truth is via joint experience. If such confirmation is not feasible, we can only \emph{estimate} the truth of claims. Establishing truth is only possible via the following mechanisms:

\begin{enumerate}[leftmargin=1.8em, itemsep=4pt, resume]
  \item \textbf{Reputation}: The history of truthfulness of the source's claims. Reliable sources are reputed to have demonstrated verifiable truth repeatedly.
  \item \textbf{Survey}: By aggregating the veracities of claims across many sources, the objective perception of claims may be estimated. This fundamentally hinges on the reasonable expectation of sources not colluding, and is very much dependent on the reputations of all sources involved.
\end{enumerate}

\begin{remark}
\begin{itemize}[leftmargin=1.2em, itemsep=3pt, topsep=2pt]
  \item By definition of truth and with all the mechanisms of truth establishment discussed above, truth is intimately related to \emph{agreement} or \emph{consensus}. Consensus, agreement, reinforcement, alignment are the modes through which shared or objective perceptions are realized.
  \item Establishing truth might not be universally pursued due to the collection of evidence being impossible, unaffordable, undesired or not considered worthwhile. However, knowledge may nevertheless be propagated without validation and/or with modifications. It is sufficient to model the truthfulness of knowledge \emph{possessed} as a limitation of perception, even if this limitation is outside the source.
\end{itemize}
\end{remark}

\begin{figure}[t]
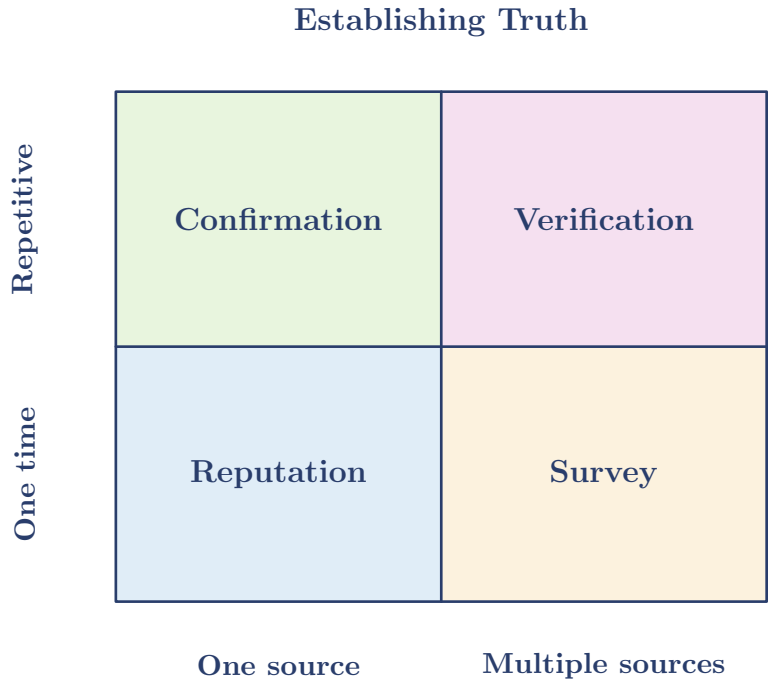

  \centering
  \begin{asy}
import math;
size(10cm, 9cm, IgnoreAspect);

pen cb        = rgb(0.16, 0.24, 0.42);
pen col_conf  = rgb(0.91, 0.96, 0.87);
pen col_verif = rgb(0.96, 0.88, 0.94);
pen col_rep   = rgb(0.88, 0.93, 0.97);
pen col_surv  = rgb(0.99, 0.95, 0.87);
pen bpen      = cb + linewidth(1.0pt);

real W = 10, H = 8, mx = W/2, my = H/2;

fill((0,my)--(mx,my)--(mx,H)--(0,H)--cycle,    col_conf);
fill((mx,my)--(W,my)--(W,H)--(mx,H)--cycle,    col_verif);
fill((0,0)--(mx,0)--(mx,my)--(0,my)--cycle,    col_rep);
fill((mx,0)--(W,0)--(W,my)--(mx,my)--cycle,    col_surv);

draw((0,0)--(W,0)--(W,H)--(0,H)--cycle, bpen);
draw((mx,0)--(mx,H), bpen);
draw((0,my)--(W,my), bpen);

label("\textbf{Confirmation}", (mx/2,    my+my/2), cb + fontsize(12pt));
label("\textbf{Verification}", (mx+mx/2, my+my/2), cb + fontsize(12pt));
label("\textbf{Reputation}",   (mx/2,    my/2),    cb + fontsize(12pt));
label("\textbf{Survey}",       (mx+mx/2, my/2),    cb + fontsize(12pt));

label(rotate(90)*"\textbf{Repetitive}", (-1.4, my+my/2), cb + fontsize(11pt));
label(rotate(90)*"\textbf{One time}",   (-1.4, my/2),    cb + fontsize(11pt));

label("\textbf{One source}",       (mx/2,    -1.0), cb + fontsize(11pt));
label("\textbf{Multiple sources}", (mx+mx/2, -1.0), cb + fontsize(11pt));

label("\textbf{Establishing Truth}", (W/2, H+1.1), cb + fontsize(13pt));
  \end{asy}
  \caption{The four mechanisms for establishing truth, organised by reproducibility
    (rows: repetitive vs.\ one-time) and number of independent sources
    (columns: one source vs.\ multiple sources).}
  \label{fig:establishing-truth}
\end{figure}

Figure~\ref{fig:establishing-truth} organises the four mechanisms in a grid by reproducibility (rows) and number of sources (columns).

\subsection{Likelihood of Objectivity as a Measure of Trust}
\label{sec:certitude}

Given the accumulation of knowledge and propagation of it, the question of whether a source of knowledge is worthy of \emph{trust} becomes a qualitative decision based on how likely the source is in providing an objective truth assessment for each specific claim it makes. The more likely the truth assessment of a given claim is reproducible, the more trustable the source is on that claim. Note that the likelihood of objectivity from a source on a specific claim might itself be unestablished, in which case it behaves as a source produces arbitrary claims, i.e., trust in that claim is minimal. The likelihood of the source's perception in aligning the objective truth to its own assessment is therefore the appropriate measure of trust.

\begin{remark}
Although we have thus far thought of truth and trust in relation to claims, the same approach applies to more general scenarios as well. For example, if the question is one of completing a \emph{task}, we can posit a corresponding claim being made by the source (the doer of the task) as ``this task is completed based on the following...'', and the likelihood of objectivity in this claim is the trust in the completion of the task.
\end{remark}

\section{A Mathematical Model for Truth and Trust}

\subsection{Claims}

A \emph{claim} $\Gamma \sim p_\Gamma$ over a space $\mathcal{N}$ is a generic statement. We consider the $\Gamma$-space $\mathcal{N}$ to be a manifold of claims that are meaningful, and may be \emph{perceived} as true, i.e., in accordance with the perceived reality, or false. For example, the space $\mathcal{N}$ can be ``the topic of fluid dynamics'', ``news items in English'', or ``proofs of mathematical statements''. A specific realization of a claim will be denoted $\gamma \in \mathcal{N}$, and all source properties such as knowledge quality, factual accuracy, and trustworthiness will be defined \emph{pointwise} for each such $\gamma$, rather than in expectation over $p_\Gamma$. This is essential: a source may perform well on average over $\mathcal{N}$ while being systematically unreliable on specific claims, and such pointwise failures must be captured by the model.

\subsection{Sources}

Claims are perceived through \emph{sources}, i.e., sources are the ``actors'' of perception. The perception of a specific claim $\gamma \in \mathcal{N}$ by source $\sigma$ is denoted $\Gamma_\sigma(\gamma)$, which can be thought of as a \emph{perturbed} form of $\gamma$, e.g., an interpretation, observation, experiment, measurement, attempt at proof, or exposition. The truth assessment of this perturbed claim by source $\sigma$ is denoted
\[
  \Theta_\sigma(\gamma) \triangleq \Theta_\sigma\bigl(\Gamma_\sigma(\gamma)\bigr) \in \{\top, \bot\}.
\]
$\Theta_\sigma(\gamma)$ is the source-defined truth estimator for the specific claim $\gamma$.

\subsubsection{Roles of Sources}

From this discussion, we arrive at two specific \emph{roles} for a source:
\begin{itemize}[leftmargin=1.8em, itemsep=3pt]
  \item A \emph{generative} role or capability to produce (create, generate, or assimilate and reproduce) claims $\Gamma_\sigma$.
  \item A \emph{discriminative} role or capability to self-establish truths $\Theta_\sigma$.
\end{itemize}
These two roles are inter-related. A source that is good only in its generative capability is unable to distinguish good from bad concepts. Conversely, being good only at the discriminative role implies good ability to discern good from bad, but not a creative ability or knowledge to generate claims.

\subsubsection{Knowledge Assimilation \& Augmentation}

The first part of source development is a process of gathering or getting exposed to claims $\gamma \in \mathcal{N}$ and being able to generate $\Gamma_\sigma(\gamma)$ from them. The nature of this generation determines whether the source is \emph{assimilative} or \emph{augmentative} --- an assimilative source reproduces $\gamma$ with possible perceptual artifacts, while an augmentative source introduces genuinely new truth-revealing information into $\Gamma_\sigma(\gamma)$.

\begin{remark}
This process brings forth the possibility of limits of perception being unbreachable: it is possible that a source is never able to produce an augmentative $\Gamma_\sigma(\gamma)$ for certain claims $\gamma$, either because the claim already reveals the truth unambiguously (leaving no room for augmentation), or because all sources hit a fundamental limit of perception. One such instance of a fundamental physical limit is \emph{Heisenberg's Uncertainty Principle}.
\end{remark}

\subsection{Interaction of Claims and Sources}

From the formulation laid down here, we are primarily concerned with:
\begin{itemize}[leftmargin=1.8em, itemsep=3pt]
  \item \textbf{Truth in claims}: The \emph{objective} measure of the truthfulness of a specific claim $\gamma$, i.e., whether $\Theta(\gamma) = \top$. This latent objective truth $\Theta(\gamma)$ is not directly observable. Since only perceptions $\Theta_\sigma(\gamma)$ are observable, we provide estimates based on these in the following sections.
  \item \textbf{Trust in sources}: The measure of objectivity in a source's perception of truth for claims $\gamma$ over a space $\mathcal{N}$. We formalize this in subsequent sections.
\end{itemize}

\subsection{Truth}

For a specific claim $\gamma \in \mathcal{N}$ perceived through $n \in \mathbb{N}$ samples $\{\Gamma_i(\gamma)\}_{i=1}^n$ from sources, we define the truth \emph{estimate} of $\gamma$ as
\[
  \hat\Theta_n(\gamma) \triangleq \bigvee_{i=1}^n \Theta_i\bigl(\Gamma_i(\gamma)\bigr),
\]
an aggregation across the $n$ perceptions of that specific claim. This definition conforms with the perception and reproducibility requirements laid down previously, and is explicitly pointwise in $\gamma$.

The exact nature of the aggregation $\bigvee$ will be discussed subsequently. Examples of this aggregation used in common practice are majority, super-majority or unanimous voting as in different voting mechanisms and democratic elections, or expert reviews established in academic publications.

\subsubsection{Truth as an Asymptote}

We desire to choose the aggregation $\bigvee$ above such that for each $\gamma \in \mathcal{N}$, the limit
\[
  \hat\Theta_n(\gamma) \xrightarrow[n\rightarrow\infty]{p} \hat\Theta(\gamma)
\]
exists, which we refer to as the \emph{objective truth} of the specific claim $\gamma$. With this, we let the truth $\Theta(\gamma)$ be \emph{approximated} by this pointwise limit, i.e., $\Theta(\gamma) \approx \hat\Theta(\gamma)$. Whereas $\Theta(\gamma)$ may not be observable, $\hat\Theta(\gamma)$ is \emph{estimable} and \emph{approximable} for each $\gamma$. Note however that $\hat\Theta(\gamma)$ may be distinct from $\Theta(\gamma)$ due to imperceptible effects or systematic errors and biases inherent to accessible perception modes. For all practical purposes however, the approximation $\Theta(\gamma) \approx \hat\Theta(\gamma)$ is the only means of understanding $\Theta(\gamma)$.

\begin{remark}
  While it is undeniable that truth is \emph{binary}, the discussions presented here are easily generalized to claims where $K \geq 2$ \emph{stances} are possible. This is often the case with ``choices'' made by sources in general, but we will continue to deal with binary truth assessments here.
\end{remark}

\subsection{A Source Model}

\begin{figure}[t]
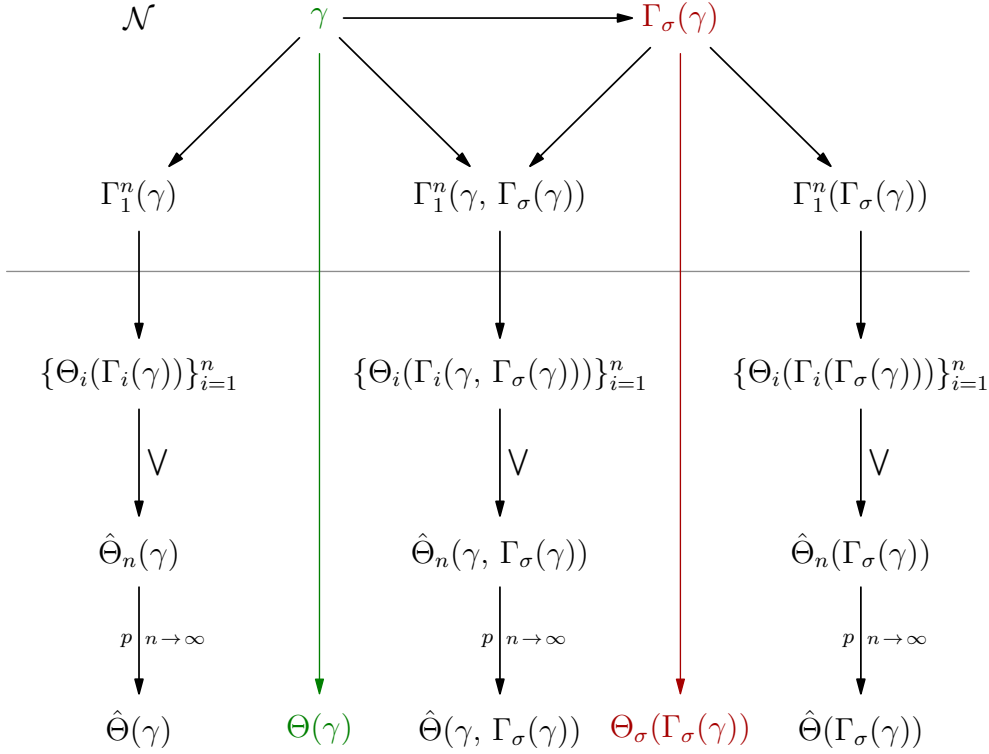

  \centering
  \begin{asy}
import math;
size(13cm, 10cm, IgnoreAspect);

// ── Pens ────────────────────────────────────────────────────────────────────
pen gpn = rgb(0.0, 0.50, 0.0);           // green  – latent/original
pen rpn = rgb(0.65, 0.0, 0.0);           // red    – source perception
pen apn = black + linewidth(0.65pt);     // arrow body
arrowbar Arr = Arrow(size=5.5pt);

// ── Node coordinates ────────────────────────────────────────────────────────
pair gam    = ( 3.0,  2.0);   // γ
pair gsig   = ( 9.0,  2.0);   // Γ_σ(γ)
pair Gng    = ( 0.0,  0.0);   // Γ_1^n(γ)
pair Gngsig = ( 6.0,  0.0);   // Γ_1^n(γ, Γ_σ(γ))  — midpoint of gam.x and gsig.x
pair Gnsig  = (12.0,  0.0);   // Γ_1^n(Γ_σ(γ))
pair Nsp    = ( 0.0,  2.0);   // 𝒩

// ── Arrow helper: trim s0 units at tail, s1 units at head ────────────────────
void darr(pair a, pair b, real s0, real s1) {
    pair u = unit(b - a);
    draw((a + s0*u) -- (b - s1*u), apn, Arr);
}

// ── Arrows (trimmed so heads/tails clear the label boxes) ────────────────────
darr(gam,  gsig,   0.40, 0.80);  // γ  →  Γ_σ(γ)
darr(gam,  Gng,    0.40, 0.60);  // γ  →  Γ_1^n(γ)
darr(gam,  Gngsig, 0.40, 0.60);  // γ  →  Γ_1^n(γ, Γ_σ(γ))
darr(gsig, Gngsig, 0.60, 0.60);  // Γ_σ(γ)  →  Γ_1^n(γ, Γ_σ(γ))
darr(gsig, Gnsig,  0.60, 0.60);  // Γ_σ(γ)  →  Γ_1^n(Γ_σ(γ))

// ── Node labels ─────────────────────────────────────────────────────────────
label("$\gamma$",
      gam,    gpn  + fontsize(12pt));
label("$\Gamma_\sigma(\gamma)$",
      gsig,   rpn  + fontsize(12pt));
label("$\Gamma_1^n(\gamma)$",
      Gng,    black + fontsize(12pt));
label("$\Gamma_1^n\!\left(\gamma,\,\Gamma_\sigma(\gamma)\right)$",
      Gngsig, black + fontsize(12pt));
label("$\Gamma_1^n\!\left(\Gamma_\sigma(\gamma)\right)$",
      Gnsig,  black + fontsize(12pt));
label("$\mathcal{N}$",
      Nsp,    black + fontsize(12pt));

// ── Horizontal boundary: 𝒩 (claim space) above / truth space below ──────────
real sep_y = -0.85;
draw((-2.2, sep_y) -- (13.8, sep_y), gray(0.55) + linewidth(0.5pt));

// ── Truth-assessment collections (below separator) ───────────────────────────
pair Tng    = ( 0.0, -2.0);   // {Θ_i(Γ_i(γ))}_{i=1}^n
pair Tngsig = ( 6.0, -2.0);   // {Θ_i(Γ_i(γ, Γ_σ(γ)))}_{i=1}^n
pair Tnsig  = (12.0, -2.0);   // {Θ_i(Γ_i(Γ_σ(γ)))}_{i=1}^n

darr(Gng,    Tng,    0.40, 0.40);   // Γ_1^n(γ)           →  {Θ_i…}
darr(Gngsig, Tngsig, 0.40, 0.40);   // Γ_1^n(γ,Γ_σ(γ))    →  {Θ_i…}
darr(Gnsig,  Tnsig,  0.40, 0.40);   // Γ_1^n(Γ_σ(γ))      →  {Θ_i…}

label("$\{\Theta_i(\Gamma_i(\gamma))\}_{i=1}^n$",
      Tng,    black + fontsize(12pt));
label("$\{\Theta_i(\Gamma_i(\gamma,\,\Gamma_\sigma(\gamma)))\}_{i=1}^n$",
      Tngsig, black + fontsize(12pt));
label("$\{\Theta_i(\Gamma_i(\Gamma_\sigma(\gamma)))\}_{i=1}^n$",
      Tnsig,  black + fontsize(12pt));

// ── \hat\Theta_n aggregates with ∨ labels on arrows ─────────────────────────
pair Hng    = ( 0.0, -4.0);   // \hat\Theta_n(γ)
pair Hngsig = ( 6.0, -4.0);   // \hat\Theta_n(γ, Γ_σ(γ))
pair Hnsig  = (12.0, -4.0);   // \hat\Theta_n(Γ_σ(γ))

void varr(pair a, pair b, real s0, real s1) {
    pair u = unit(b - a);
    draw((a + s0*u) -- (b - s1*u), apn, Arr);
    label("$\bigvee$", (a + b)/2, E, fontsize(9pt));
}

varr(Tng,    Hng,    0.40, 0.40);   // {Θ_i(γ)}  →  \hat\Theta_n(γ)
varr(Tngsig, Hngsig, 0.40, 0.40);   // {Θ_i(γ,Γ_σ(γ))}  →  \hat\Theta_n(γ,Γ_σ(γ))
varr(Tnsig,  Hnsig,  0.40, 0.40);   // {Θ_i(Γ_σ(γ))}  →  \hat\Theta_n(Γ_σ(γ))

label("$\hat\Theta_n(\gamma)$",
      Hng,    black + fontsize(12pt));
label("$\hat\Theta_n(\gamma,\,\Gamma_\sigma(\gamma))$",
      Hngsig, black + fontsize(12pt));
label("$\hat\Theta_n(\Gamma_\sigma(\gamma))$",
      Hnsig,  black + fontsize(12pt));

// ── Asymptotic truth limits (n → ∞) ─────────────────────────────────────────
pair Lng    = ( 0.0, -6.0);   // \hat\Theta(γ)
pair Lngsig = ( 6.0, -6.0);   // \hat\Theta(γ, Γ_σ(γ))
pair Lnsig  = (12.0, -6.0);   // \hat\Theta(Γ_σ(γ))

void parr(pair a, pair b, real s0, real s1) {
    pair u = unit(b - a);
    pair mid = (a + b)/2;
    draw((a + s0*u) -- (b - s1*u), apn, Arr);
    label("$p$",              mid, W, fontsize(8pt));
    label("$n\!\to\!\infty$", mid, E, fontsize(7pt));
}

parr(Hng,    Lng,    0.40, 0.40);   // \hat\Theta_n(γ)  →  \hat\Theta(γ)
parr(Hngsig, Lngsig, 0.40, 0.40);   // \hat\Theta_n(γ,Γ_σ(γ))  →  \hat\Theta(γ,Γ_σ(γ))
parr(Hnsig,  Lnsig,  0.40, 0.40);   // \hat\Theta_n(Γ_σ(γ))  →  \hat\Theta(Γ_σ(γ))

label("$\hat\Theta(\gamma)$",
      Lng,    black + fontsize(12pt));
label("$\hat\Theta(\gamma,\,\Gamma_\sigma(\gamma))$",
      Lngsig, black + fontsize(12pt));
label("$\hat\Theta(\Gamma_\sigma(\gamma))$",
      Lnsig,  black + fontsize(12pt));

// ── Latent truth Θ(γ) and subjective truth Θ_σ(Γ_σ(γ)) ─────────────────────
pair ThetaL = ( 3.0, -6.0);   // Θ(γ)               — latent, green
pair ThetaS = ( 9.0, -6.0);   // Θ_σ(Γ_σ(γ))        — subjective, red

pen dap = linewidth(0.65pt) + linetype("4 3");   // dashed style

// γ  ··>  Θ(γ)   (long vertical dashed, green)
draw((gam  + 0.40*(0,-1)) -- (ThetaL - 0.40*(0,-1)), gpn, Arr);
// Γ_σ(γ)  ··>  Θ_σ(Γ_σ(γ))   (long vertical dashed, red)
draw((gsig + 0.40*(0,-1)) -- (ThetaS - 0.40*(0,-1)), rpn, Arr);

label("$\Theta(\gamma)$",
      ThetaL, gpn + fontsize(12pt));
label("$\Theta_\sigma(\Gamma_\sigma(\gamma))$",
      ThetaS, rpn + fontsize(12pt));
  \end{asy}
  \caption{Full source-model chain for a specific claim $\gamma$.
    \textit{Claim space ($\mathcal{N}$, above the separator):} The latent claim $\gamma$ (green) is mapped by source $\sigma$ to its perception $\Gamma_\sigma(\gamma)$ (red); together they generate three collections of samples from other sources --- direct perceptions $\Gamma_1^n(\gamma)$, the joint perceptions $\Gamma_1^n(\gamma,\Gamma_\sigma(\gamma))$, and derived perceptions $\Gamma_1^n(\Gamma_\sigma(\gamma))$.
    \textit{Truth-assessment layer (below separator):} Each collection yields individual assessments $\{\Theta_i(\Gamma_i(\cdot))\}_{i=1}^n$, aggregated via $\bigvee$ into finite-sample estimates $\hat\Theta_n(\cdot)$.
    \textit{Asymptotic layer:} As $n\to\infty$ (in probability), these converge to the objective truths $\hat\Theta(\gamma)$, $\hat\Theta(\gamma,\Gamma_\sigma(\gamma))$, and $\hat\Theta(\Gamma_\sigma(\gamma))$. The latent truth $\Theta(\gamma)$ (green, solid arrow from $\gamma$) and the source's subjective assessment $\Theta_\sigma(\Gamma_\sigma(\gamma))$ (red, solid arrow from $\Gamma_\sigma(\gamma)$) are shown alongside, as the targets that the estimation chain approximates.}
  \label{fig:claim-space}
\end{figure}

In order to model a source over the space $\mathcal{N}$, we consider the full chain of dependencies for each specific claim $\gamma \in \mathcal{N}$. Figure~\ref{fig:claim-space} illustrates this chain end-to-end: from the claim-space (above the separator) through the truth-assessment collections and finite-sample aggregates, down to the asymptotic objective truth limits and the two distinguished special cases of the latent $\Theta(\gamma)$ and the source-specific $\Theta_\sigma(\Gamma_\sigma(\gamma))$.

\noindent The figure traces the complete inference chain for a specific $\gamma \in \mathcal{N}$.
In the claim space (above the separator), $\gamma$ (green) is the latent original claim and
$\Gamma_\sigma(\gamma)$ (red) is the source's perception; together they fan out into three sample collections.
Below the separator, those collections produce individual truth assessments $\{\Theta_i\}$, which are
aggregated via $\bigvee$ into finite-sample estimates $\hat\Theta_n$, converging in probability to the
asymptotic objective truths $\hat\Theta$ as $n\to\infty$.
The green and red vertical arrows identify the two special quantities against which the chain is
benchmarked: the latent $\Theta(\gamma)$ (the unobservable ground truth) and the source's own
assessment $\Theta_\sigma(\Gamma_\sigma(\gamma))$ (the subjective single-source reading).
Every quantity is defined pointwise for the specific $\gamma$.

\subsubsection{Truth Notions and Interactions}
\label{sec:truth-notions}

\begin{figure}[t]
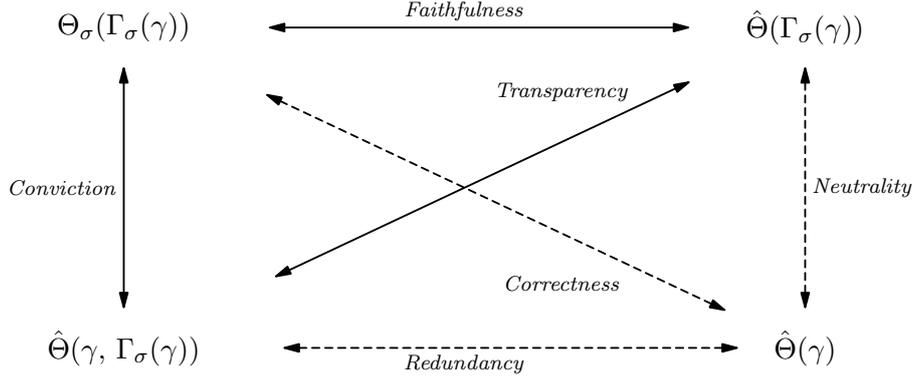

  \centering
  \begin{asy}
import math;
size(12cm, 5cm, IgnoreAspect);

pen apn  = black + linewidth(0.7pt);
pen dpn  = black + linewidth(0.7pt) + linetype("4 3");
arrowbar A2 = Arrows(size=5.5pt);

pair TL = (2, 3);   // Θ_σ(Γ_σ(γ))
pair TR = (9, 3);   // \hat\Theta(Γ_σ(γ))
pair BL = (2, 0);   // \hat\Theta(γ,Γ_σ(γ))
pair BR = (9, 0);   // \hat\Theta(γ)

void aarr(pair a, pair b, real s0, real s1, pen p) {
    pair u = unit(b - a);
    draw((a + s0*u) -- (b - s1*u), p, A2);
}

// ── Solid: triangle TL–TR–BL ────────────────────────────────────────────
aarr(TL, TR, 1.50, 1.20, apn);   // Faithfulness  (top horizontal)
aarr(TL, BL, 0.38, 0.38, apn);   // Conviction    (left vertical)
aarr(TR, BL, 1.30, 1.70, apn);   // Transparency  (diagonal)

// ── Dashed: three edges to BR ───────────────────────────────────────────
aarr(TL, BR, 1.60, 0.90, dpn);   // Correctness   (diagonal)
aarr(TR, BR, 0.38, 0.38, dpn);   // Neutrality    (right vertical)
aarr(BL, BR, 1.65, 0.70, dpn);   // Redundancy    (bottom horizontal)

// ── Node labels ─────────────────────────────────────────────────────────
label("$\Theta_\sigma(\Gamma_\sigma(\gamma))$",
      TL, black + fontsize(11pt));
label("$\hat\Theta(\Gamma_\sigma(\gamma))$",
      TR, black + fontsize(11pt));
label("$\hat\Theta(\gamma,\,\Gamma_\sigma(\gamma))$",
      BL, black + fontsize(11pt));
label("$\hat\Theta(\gamma)$",
      BR, black + fontsize(11pt));

// ── Arrow labels ────────────────────────────────────────────────────────
label("\textit{Faithfulness}", (TL+TR)/2,        N,   fontsize(8pt));
label("\textit{Conviction}",   (TL+BL)/2,        W,   fontsize(8pt));
label("\textit{Transparency}", (6.5, 2.4),             fontsize(8pt));
label("\textit{Correctness}",  (6.5, 0.6),             fontsize(8pt));
label("\textit{Neutrality}",   (TR+BR)/2,        E,   fontsize(8pt));
label("\textit{Redundancy}",   (BL+BR)/2,        S,   fontsize(8pt));
  \end{asy}
  \caption{The four notions of truth and their six bilateral interactions.
    \textit{Solid arrows}: interactions among the three notions that exclude $\hat\Theta(\gamma)$
    (Faithfulness, Conviction, Transparency).
    \textit{Dashed arrows}: interactions that directly involve $\hat\Theta(\gamma)$
    (Correctness, Neutrality, Redundancy).}
  \label{fig:truth-notions}
\end{figure}

Based on the discussion in the previous section, the four notions of truth and all six bilateral interactions between them are depicted in Figure~\ref{fig:truth-notions}. Solid arrows connect the three notions that exclude $\hat\Theta(\gamma)$ outright; dashed arrows involve $\hat\Theta(\gamma)$ directly.
\noindent The interactions between these are as below.

\begin{itemize}[leftmargin=1.8em, itemsep=5pt]
  \item \textbf{Faithfulness}: The source's stance aligns with the objective stance based on its perception,
  $\Theta_\sigma(\Gamma_\sigma(\gamma)) = \hat\Theta(\Gamma_\sigma(\gamma))$.

  \item \textbf{Conviction}: The source's stance is vindicated by joint consensus,
  $\Theta_\sigma(\Gamma_\sigma(\gamma)) = \hat\Theta(\gamma, \Gamma_\sigma(\gamma))$.

  \item \textbf{Transparency}: The source's perception is sufficient for objective truth assessment and disclosing the original claim $\gamma$ adds nothing,
  $\hat\Theta(\Gamma_\sigma(\gamma)) = \hat\Theta(\gamma, \Gamma_\sigma(\gamma))$.

  \item \textbf{Correctness}: The source's stance agrees with the objective consensus on the original claim,
  $\Theta_\sigma(\Gamma_\sigma(\gamma)) = \hat\Theta(\gamma)$.

  \item \textbf{Neutrality}: The source's perception doesn't shift consensus,
  $\hat\Theta(\Gamma_\sigma(\gamma)) = \hat\Theta(\gamma)$.

  \item \textbf{Redundancy}: The source's perception doesn't add to the original claim,
  $\hat\Theta(\gamma, \Gamma_\sigma(\gamma)) = \hat\Theta(\gamma)$.
\end{itemize}

The first three interactions of faithfulness, conviction, and transparency are unconditional desiderata: a reliable source should satisfy all three, and approximately achieving any two approximately implies the third. The remaining three --- correctness, neutrality, and redundancy --- form a qualitatively different cluster. Together they express the condition that the source's perception leaves the objective truth estimate unchanged: the source neither shifts consensus nor adds information beyond what $\gamma$ already contains. This is precisely the \emph{assimilative regime}.

Formally, a source is said to be in the \emph{assimilative regime} for claim $\gamma$ when redundancy holds approximately, i.e., $\hat\Theta(\gamma, \Gamma_\sigma(\gamma)) \approx \hat\Theta(\gamma)$. Conversely, the source is in the \emph{augmentative regime} when redundancy is violated, i.e., when $\Gamma_\sigma(\gamma)$ carries truth-revealing information not already present in $\gamma$ itself. In cases of new discoveries, departing from correctness, neutrality and redundancy is not only acceptable but expected of a genuinely creative source; what matters is that faithfulness, conviction and transparency are maintained throughout.

\begin{remark}
It is worth noting that the failure of conviction to entail faithfulness and transparency is the \emph{atypical} case. When $\Gamma_\sigma(\gamma)$ is a \emph{complete} perception --- one that stands by itself as a substantive, self-contained claim --- then $\hat\Theta(\Gamma_\sigma(\gamma)) \approx \hat\Theta(\gamma, \Gamma_\sigma(\gamma))$ holds naturally, since other sources assessing $\Gamma_\sigma(\gamma)$ in isolation reach the same consensus as those who also have $\gamma$. In this case, conviction pulls faithfulness and transparency along with it, and the three desiderata collapse into one.

The pathological case arises when $\Gamma_\sigma(\gamma)$ is an \emph{incomplete} perception --- one that does not stand by itself and requires $\gamma$ to become substantive. Such a perception is more aptly described as a \emph{relation} to $\gamma$ than a perception in its own right: think of a proof step that presupposes earlier definitions, a measurement whose interpretation depends on the experimental setup, or a comparative judgment (``this is an improvement'') that is vacuous without its referent. In all such cases, sources evaluating $\Gamma_\sigma(\gamma)$ alone lack the necessary context to form a reliable consensus, so $\hat\Theta(\Gamma_\sigma(\gamma))$ diverges from $\hat\Theta(\gamma, \Gamma_\sigma(\gamma))$, transparency breaks, and with it faithfulness. The source may still have conviction, but only for observers who already possess $\gamma$ --- a considerably weaker guarantee.

This gives a practical criterion: for conviction to entail the full triad, it is sufficient that the source's perceptions be \emph{complete} --- communicable and assessable as standalone claims, without reference to their origin.
\end{remark}

\subsubsection{The Influence of Perception on Objective Truth}

\begin{figure}[t]
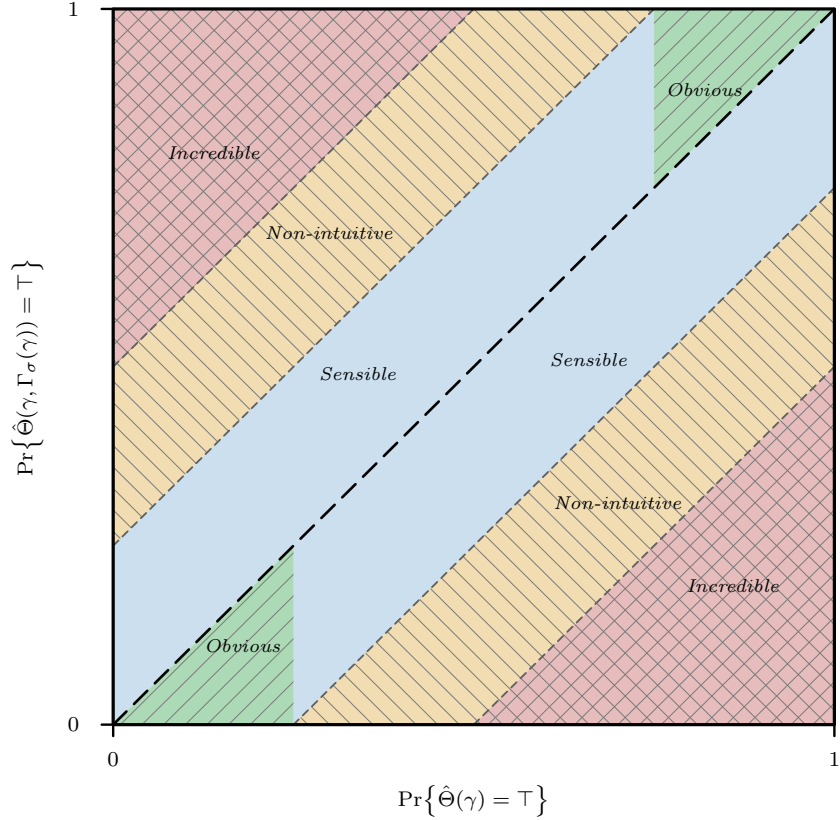

  \centering
  \begin{asy}
import math;
size(11cm, 11cm);

real S = 10;

pen col_obvious  = rgb(0.38, 0.72, 0.45) + opacity(0.52);
pen col_sensible = rgb(0.55, 0.72, 0.85) + opacity(0.45);
pen col_nonint   = rgb(0.88, 0.70, 0.32) + opacity(0.45);
pen col_incred   = rgb(0.78, 0.38, 0.38) + opacity(0.42);

pen hatch_pen  = gray(0.48) + linewidth(0.35);
pen border_pen = black + linewidth(1.2);
pen diag_pen   = black + linewidth(1.0) + linetype("7 4");
pen off_pen    = gray(0.38) + linewidth(0.7) + linetype("4 3");
pen axis_pen   = black + linewidth(0.9);

pair P(real x, real y) { return (x*S, y*S); }

void hatchNE(path poly) {
    picture htmp;
    real sp = 0.30;
    for (real d = -2*S; d <= 3*S; d += sp)
        draw(htmp, (d, 0)--(d+S, S), hatch_pen);
    clip(htmp, poly);
    add(htmp);
}

void hatchNW(path poly) {
    picture htmp;
    real sp = 0.30;
    for (real d = -S; d <= 3*S; d += sp)
        draw(htmp, (d, S)--(d+S, 0), hatch_pen);
    clip(htmp, poly);
    add(htmp);
}

void hatchX(path poly) { hatchNE(poly); hatchNW(poly); }

void fpoly(path poly, pen p) { fill(poly, p); }

path inc_up  = P(0,0.5)--P(0,1)--P(0.5,1)--cycle;
path inc_dn  = P(0.5,0)--P(1,0)--P(1,0.5)--cycle;
path noni_up = P(0,0.25)--P(0,0.5)--P(0.5,1)--P(0.75,1)--cycle;
path noni_dn = P(0.25,0)--P(0.5,0)--P(1,0.5)--P(1,0.75)--cycle;
path sens_up = P(0,0)--P(0,0.25)--P(0.75,1)--P(0.75,0.75)--cycle;
path sens_dn = P(0.25,0.25)--P(0.25,0)--P(1,0.75)--P(1,1)--cycle;
path obv_up  = P(0.75,0.75)--P(0.75,1)--P(1,1)--cycle;
path obv_dn  = P(0,0)--P(0.25,0)--P(0.25,0.25)--cycle;

fpoly(inc_up, col_incred);   fpoly(inc_dn, col_incred);
fpoly(noni_up, col_nonint);  fpoly(noni_dn, col_nonint);
fpoly(sens_up, col_sensible); fpoly(sens_dn, col_sensible);
fpoly(obv_up, col_obvious);  fpoly(obv_dn, col_obvious);

hatchX(inc_up);   hatchX(inc_dn);
hatchNW(noni_up); hatchNW(noni_dn);
hatchNE(obv_up);  hatchNE(obv_dn);

draw(P(0,0)--P(1,1),       diag_pen);
draw(P(0,0.25)--P(0.75,1), off_pen);
draw(P(0,0.5)--P(0.5,1),   off_pen);
draw(P(0.25,0)--P(1,0.75), off_pen);
draw(P(0.5,0)--P(1,0.5),   off_pen);

draw(box(P(0,0),P(1,1)), border_pen);

real tl = 0.16;
draw(P(0,0)--P(0,0)-(0,tl), axis_pen);
draw(P(1,0)--P(1,0)-(0,tl), axis_pen);
label("$0$", P(0,0)-(0,0.48), fontsize(8pt));
label("$1$", P(1,0)-(0,0.48), fontsize(8pt));
draw(P(0,0)--P(0,0)-(tl,0), axis_pen);
draw(P(0,1)--P(0,1)-(tl,0), axis_pen);
label("$0$", P(0,0)-(0.55,0), fontsize(8pt));
label("$1$", P(0,1)-(0.55,0), fontsize(8pt));

label("$\Pr\bigl\{\hat\Theta(\gamma)=\top\bigr\}$",
      P(0.5,0)-(0,1.05), fontsize(8pt));
label(rotate(90)*"$\Pr\bigl\{\hat\Theta(\gamma,\Gamma_\sigma(\gamma))=\top\bigr\}$",
      P(0,0.5)-(1.20,0), fontsize(8pt));

label("\textit{Obvious}",        P(0.820, 0.888), fontsize(7pt));
label("\textit{Obvious}",        P(0.180, 0.110), fontsize(7pt));
label("\textit{Incredible}",     P(0.140, 0.800), fontsize(7pt));
label("\textit{Incredible}",     P(0.860, 0.195), fontsize(7pt));
label("\textit{Non-intuitive}",  P(0.300, 0.688), fontsize(7pt));
label("\textit{Non-intuitive}",  P(0.700, 0.310), fontsize(7pt));
label("\textit{Sensible}",       P(0.340, 0.490), fontsize(7pt));
label("\textit{Sensible}",       P(0.660, 0.510), fontsize(7pt));
  \end{asy}
  \caption{Classification of source behaviour by the objective truth of the original claim $\hat\Theta(\gamma)$ (horizontal axis) versus the joint objective truth after incorporating the source's perception $\hat\Theta(\gamma, \Gamma_\sigma(\gamma))$ (vertical axis). The dashed diagonal is the line of no influence. Parallel offset lines delimit the four regions.}
  \label{fig:conviction-plot}
\end{figure}

The relationship between $\hat{\Theta}(\gamma)$ and $\hat{\Theta}(\gamma, \Gamma_\sigma(\gamma))$ is illustrated in Figure~\ref{fig:conviction-plot}. The four regions correspond to different qualitative regimes of source behaviour:

\begin{itemize}[leftmargin=1.8em, itemsep=4pt]
  \item \textbf{Obvious}: The source's perception closely tracks the original claim's truth value; no significant shift occurs and the prior objective stance is reinforced.
  \item \textbf{Sensible}: The source introduces a moderate, graded shift in the objective truth estimate --- the perception is informative but not transformative, either reinforcing or mildly refuting the prior objective stance.
  \item \textbf{Non-intuitive}: The source's perception substantially overturns the prior consensus; its contribution is strongly augmentative.
  \item \textbf{Incredible}: The source's perception is so far from the original claim's truth that it could only be trusted in the augmentative regime with very high conviction.
\end{itemize}

The four regions characterize the qualitative relationship between prior and posterior objective truth. The formal metrics quantifying a source's behavior within these regions are introduced in Section~\ref{sec:desiderata-formal}.

\subsubsection{Estimating Truth}
\label{sec:est-truth-math}

Tying this mathematical model back to the methods of establishing truth described in Section~\ref{sec:establishing-truth} is illuminating:
\begin{itemize}[leftmargin=1.8em, itemsep=4pt]
  \item When we are reliant on a single source's one-time perception, the truth is based on $\Theta_\sigma(\Gamma_\sigma(\gamma))$ --- the only way to be confident of this being true is via the source's \textbf{reputation} built over the course of time.
  \item When we can rely on many sources' perceptions of a one-time event, the truth is established via $\hat\Theta(\gamma)$, i.e., the \textbf{survey} approach.
  \item When we rely on repeated reproduction via the same experiment, the truth is based on $\hat\Theta(\Gamma_\sigma(\gamma))$, \textbf{confirmation} of the objective stance on the perception through the experiment.
  \item When we battle-test the possibilities via repeated testing and verification across many sources, the truth is based on $\hat\Theta(\gamma, \Gamma_\sigma(\gamma))$, born out of \textbf{verification} via independent analysis and peer review.
\end{itemize}

\subsubsection{Desiderata for Trust}
\label{sec:reputation}
\label{sec:desiderata-formal}

The interactions identified in Section~\ref{sec:truth-notions} are binary conditions. We now define their probabilistic counterparts, which form the basis for measuring source reliability in practice. We refer to the probability of conviction as the \emph{conviction} of source $\sigma$ for claim $\gamma$
\[
  C_\sigma(\gamma) \triangleq \Pr\bigl\{\Theta_\sigma(\Gamma_\sigma(\gamma)) = \hat\Theta(\gamma, \Gamma_\sigma(\gamma))\bigr\} \;\in\; [0, 1].
\]
Conviction is the operative mechanism through which the likelihood of objectivity introduced in Section~\ref{sec:certitude} is formalized: a source with high conviction is one whose stances are reliably vindicated by independent consensus.

\begin{remark}
  Two further quantities relate conviction to the other truth notions. \emph{Persuasiveness} measures the excess tendency of a source's stance to align with the objective truth of its own perception over the objective truth of the original claim
\[
  P_\sigma(\gamma) \triangleq \Pr\bigl\{\Theta_\sigma(\Gamma_\sigma(\gamma)) = \hat\Theta(\Gamma_\sigma(\gamma))\bigr\} - \Pr\bigl\{\Theta_\sigma(\Gamma_\sigma(\gamma)) = \hat\Theta(\gamma)\bigr\} \;\in\; [-1, 1],
\]
and \emph{demonstrability} measures the excess tendency to align with the joint objective truth over the original
\[
  D_\sigma(\gamma) \triangleq \Pr\bigl\{\Theta_\sigma(\Gamma_\sigma(\gamma)) = \hat\Theta(\gamma, \Gamma_\sigma(\gamma))\bigr\} - \Pr\bigl\{\Theta_\sigma(\Gamma_\sigma(\gamma)) = \hat\Theta(\gamma)\bigr\} \;\in\; [-1, 1].
\]
High persuasiveness with low demonstrability indicates that the persuasion is only effective when the original claim is concealed or unobservable --- a situation reflecting lack of conviction and transparency. Demonstrability is the metric used to declare winners in Oxford-style debates: the correctness term $\Pr\{\Theta_\sigma(\Gamma_\sigma(\gamma)) = \hat\Theta(\gamma)\}$ denotes votes in favor of the source's stance prior to the debate, and the conviction term $\Pr\{\Theta_\sigma(\Gamma_\sigma(\gamma)) = \hat\Theta(\gamma, \Gamma_\sigma(\gamma))\}$ denotes votes after. Demonstrability therefore quantifies the \emph{sway} of opinion toward the source's stance, and the winner is the participant with the largest such sway.
\end{remark}

While the approaches of survey, confirmation and verification allow for consultation of objective notions of truth, i.e., variants of the asymptote $\hat\Theta$, the natural question that remains is what constitutes the most principled approach to establishing the \emph{reputation} of a source that can be relied upon as a measure of trust when only $\Theta_\sigma(\Gamma_\sigma(\gamma))$ is available. Based on prior discussion, we have the ability to base reputation on any of \emph{conviction}, \emph{correctness} or \emph{faithfulness}.

We have previously explained that correctness is desirable only in the assimilative regime. In principle, whether we are in an augmentative regime is inestimable \emph{a priori}, for if it were, the augmentation is already estimated and, by definition, not augmentation. It is therefore preferred to base the reputation of a source based on its conviction and faithfulness. If there were a single test feasible, it is preferred to be a test of conviction. A test of correctness would penalize innovation, while a test of faithfulness could compound biases of perception --- a systematically biased source may be faithful to its own skewed perception while diverging from objective consensus.

\begin{remark}
Even in well-established areas, and even without the underlying truth being perturbed, genuine creativity and innovation in a source's perceptions are very much possible and desirable. These operational improvements and efficiencies are, however, orthogonal to truthfulness and trust, and thus outside the scope of our framework.
\end{remark}
\begin{remark}
Apart from the benefits of leveraging conviction we have discussed so far, there is yet another advantage. While the joint perception over $(\gamma, \Gamma_\sigma(\gamma))$ may be more verbose, the actual act of truth assessment becomes one of verifying the proof, as opposed to requiring reproving claims as with either faithfulness or correctness checks. This can be substantially more efficient computationally as well as more economical in practice.
\end{remark}

As noted before, conviction of the source is most often the sufficient target, with the exception of insufficiency of the source's perceptions by themselves. Optimizing for conviction also has an expectation for a source that is very relatable: 
\begin{note}
\emph{Assume} both the original claim and the source's perception can be reviewed as desired, and attempt to produce a perception such that \emph{any} such reviewer will gravitate to the source's truth estimate.
\end{note}
This is a very \emph{human} expectation --- performing tasks with an acknowledgment of them being reviewed and producing work that can be stood behind when challenged forms the basis of earning reputation in any human endeavor. Social constraints have proven that this is the most robust approach to engendering reliability and trust in systems whose participants are themselves error-prone.

We can therefore posit the following requirements for a trustable source in a trusted ecosystem:
\begin{enumerate}
  \item Facilitate reviews by being transparent and producing self-sufficient perceptions to align any reviewers with internal truth assessments.
  \item Require reputation building via such review mechanisms to establish credibility in claimed areas of expertise.
\end{enumerate}
We will elaborate on appropriate metrics for reputation next.

\subsection{Metrics for Reputation}

Having established conviction as the preferred basis for reputation in Section~\ref{sec:reputation}, we now formalize the metrics through which it is operationalized. The central challenge is that reputation must reflect not only whether a source shows conviction, but also the weight that each claim $\gamma$ deserves in the accumulation of that reputation. We argue that this weight is determined by the certitude of the objective truth --- both prior to and after the source's perception --- and that the sign of each contribution is determined by the alignment of the source's stance with the posterior objective truth.

\subsubsection{Claim Weight via Certitude of Objectivity}

For a claim $\gamma \in \mathcal{N}$, the prior certitude of objectivity is the certainty with which the objective truth of $\gamma$ is established before any source perception is introduced
\[
    w^-(\gamma) \triangleq 1 - H\big(\Pr\{\hat{\Theta}(\gamma) = \top\}\big),
\]
where $H(\cdot)$ denotes binary entropy. This weight approaches 1 when prior objectivity is certain and approaches 0 when the claim is genuinely contested. Similarly, the posterior certitude of objectivity reflects the certainty of objectivity after the source's perception $\Gamma_\sigma(\gamma)$ is included
\[
    w^+(\gamma, \sigma) \triangleq 1 - H\big(\Pr\{\hat{\Theta}(\gamma, \Gamma_\sigma(\gamma)) = \top\}\big).
\]
The joint claim weight is the product of prior and posterior certitudes:
\[
    w(\gamma, \sigma) \triangleq w^-(\gamma) \cdot w^+(\gamma, \sigma).
\]
This product weights contributions most heavily when both prior and posterior objectivity are certain, and discounts when either is uncertain. Crucially, a source that destabilizes an established consensus is naturally discounted on the posterior factor, withholding reputation credit until the claim resolves.

\subsubsection{Signed Conviction}
The direction of a source's reputation contribution is determined by whether its stance aligns with or opposes the posterior objectivity. We define the signed conviction of source $\sigma$ on claim $\gamma$ as
\[
    \tilde{C}_\sigma(\gamma) \triangleq 2 C_\sigma(\gamma) - 1 \in [-1, +1],
\]
where $C_\sigma(\gamma) = \Pr\{\Theta_\sigma(\Gamma_\sigma(\gamma)) = \hat{\Theta}(\gamma, \Gamma_\sigma(\gamma))\}$ is the conviction as defined previously. 

The signed conviction has the following natural properties
\begin{itemize}
    \item $\tilde{C}_\sigma(\gamma) \to +1$ when the source is consistently vindicated by posterior consensus and provides maximum positive contribution
    \item $\tilde{C}_\sigma(\gamma) = 0$ when the source is vindicated exactly half the time and any contribution to reputation is withheld
    \item $\tilde{C}_\sigma(\gamma) \to -1$ when the source is consistently opposed by posterior consensus and provides maximum negative contribution
\end{itemize}

\subsubsection{Reputation}
The reputation of source $\sigma$ over a realm $\mathcal{R} \subseteq \mathcal{N}$ is defined as the expected weighted signed conviction over claims in that realm
\[
    R_\sigma(\mathcal{R}) \triangleq \mathbb{E}_{\gamma \sim p_\Gamma(\cdot | \mathcal{R})}\Big[\tilde{C}_\sigma(\gamma) \cdot w(\gamma, \sigma)\Big],
\]
may be approximated based on a corpus of claims $\mathcal{R}^\prime \subseteq \mathcal{R}$ as
\begin{align}
  R_\sigma(\mathcal{R}) &\approx \frac{1}{|\mathcal{R}^\prime|}\sum_{\gamma\in\mathcal{R}^\prime} \big(\mathds{1}\{\Theta_\sigma(\Gamma_\sigma(\gamma)) = \hat{\Theta}(\gamma, \Gamma_\sigma(\gamma))\} \notag \\
  &\qquad\qquad\qquad - \mathds{1}\{\Theta_\sigma(\Gamma_\sigma(\gamma)) \neq \hat{\Theta}(\gamma, \Gamma_\sigma(\gamma))\}\big) \cdot w_{\mathcal{R}^\prime}(\gamma, \sigma),\notag
\end{align}
where $w_{\mathcal{R}^\prime}(\gamma, \sigma)$ is the estimated certitude weight over the corpus $\mathcal{R}^\prime$.

This definition has several desirable properties
\begin{enumerate}
    \item \textbf{Boundedness}: $R_\sigma(\mathcal{R}) \in [-1, +1]$, with $+1$ corresponding to a perfectly reliable source and $-1$ to a perfectly misleading one.
    \item \textbf{Claim-sensitivity}: Claims on which objectivity is uncertain contribute less to reputation, naturally disentangling source reliability from claim contentiousness.
    \item \textbf{Regime-independence}: The formulation is meaningful in both the assimilative and augmentative regimes.
    \item \textbf{Continuity}: Reputation accrues gradually as conviction and objectivity certainty accumulate, and continuous improvements compound rewards.
\end{enumerate}

\begin{remark}
The choice of aggregation operator $\bigvee$ determines convergence properties of the objective truth. Two natural families of instantiations are worth noting. In a reputation-weighted voting approach, each verifier's assessment $\Theta_i(\Gamma_i(\gamma))$ is weighted by their reputation $R_i$, creating a mutually reinforcing structure where reputation feeds back into truth estimation and truth estimation feeds forward into reputation. In a Bayesian approach, each verifier perception $\Gamma_i(\gamma)$ continuously refines the objective consensus, with the prior certitude $w^-(\gamma)$ updated sequentially as perceptions accumulate.
\end{remark}

\subsubsection{Reputation Across Source-Claim Regimes}
The behavior of $R_\sigma(\mathcal{R})$ across the source-claim regimes identified in Figure~\ref{fig:conviction-plot} is summarized in Table~\ref{tab:regimes}. The regions of Figure~\ref{fig:conviction-plot} characterize the relationship between prior and posterior consensus, while conviction determines the sign and magnitude of the source's contribution within each region.

\renewcommand{\arraystretch}{1.5}
\begin{longtable}{p{2cm} p{2cm} p{2.5cm} p{6.5cm}}
\caption{Source reputation contributions across the source-claim regimes of Figure~\ref{fig:conviction-plot}, characterized by region and signed conviction $\tilde{C}_\sigma(\gamma)$. Discounting reflects the joint claim weight $w(\gamma, \sigma)$. Withheld contributions arise naturally when signed conviction approaches zero, implementing suspension of reputation judgment on contentious or unresolved claims without requiring explicit thresholding.}
\label{tab:regimes}\\
\toprule
\textbf{Region} & \textbf{Signed} & \textbf{Reputation} & \textbf{Interpretation} \\
& \textbf{Conviction} & \textbf{Contribution} & \\
\midrule
\endfirsthead
\multicolumn{4}{c}{\tablename~\thetable{} \textit{(continued)}}\\
\toprule
\textbf{Region} & \textbf{Signed} & \textbf{Reputation} & \textbf{Interpretation} \\
& \textbf{Conviction} & \textbf{Contribution} & \\
\midrule
\endhead
\midrule
\multicolumn{4}{r}{\textit{continued on next page}}\\
\endfoot
\bottomrule
\endlastfoot
Obvious & High ($\to +1$) & Strong positive & Reinforcing assimilative source: consistently confirms established truth, the backbone of reliable knowledge \\
Obvious & Medium ($\approx 0$) & Withheld & Doubt introducer: intermittently disagrees with settled consensus without augmentative force; withheld pending stabilization but trending negative \\
Obvious & Low ($\to -1$) & Strong negative & Nonconformist: consistently opposes settled consensus without the augmentative force to move it; unambiguous reputation penalty \\
Sensible & High ($\to +1$) & Positive, discounted & Selective augmentative contributor: intervenes on partially settled claims and is consistently vindicated; discounted by posterior uncertainty \\
Sensible & Medium ($\approx 0$) & Withheld & Feather-ruffler: source perception moves partially settled consensus without clear vindication; claim contentiousness and source reliability are conflated; judgment suspended \\
Sensible & Low ($\to -1$) & Negative, discounted & Selective destabilizer: consistently wrong when augmentative on partially settled claims; discounted by posterior uncertainty \\
Non-intuitive & High ($\to +1$) & Strong positive, heavily discounted initially & Genuine innovator: dramatically moves consensus and is vindicated; paradigm-shifter or theory-establisher depending on prior truth; contribution accrues as conviction stabilizes \\
Non-intuitive & Low ($\to -1$) & Strong negative, heavily discounted initially & Inadvertent contributor: dramatically moves consensus away from own stance; contribution accrues as low conviction stabilizes \\
Incredible & High ($\to +1$) & Maximum positive, most heavily discounted initially & Paradigm-defining innovator: maximally moves consensus and is vindicated; highest reputation contribution when resolved; slowest to accrue \\
Incredible & Low ($\to -1$) & Maximum negative, most heavily discounted initially & Consequential inadvertent contributor: maximally moves consensus against own stance; strongest reputation penalty when resolved \\
\end{longtable}

\begin{remark}
The continuity property of reputation has a direct implication for the assessment of source reputation in practice. Since reputation accrues gradually through the accumulation of weighted conviction across claims, a single observation of a source's behavior is insufficient to characterize its reliability. This is particularly acute in the non-intuitive and incredible regions, where the transient period before conviction stabilizes may be extended, and where point-in-time assessment may capture a source mid-trajectory and mischaracterize its reliability. Continuous observation of conviction and consensus trajectories is therefore not merely a practical convenience but a theoretical necessity for accurate reputation assessment.

A claim $\gamma$ is intrinsically contentious when $w^-(\gamma) \approx 0$, i.e., when prior consensus is maximally uncertain. In this regime, the joint weight $w(\gamma, \sigma)$ approaches zero regardless of the source's conviction, naturally suspending reputation judgment. This disentangles source reliability from claim contentiousness without requiring explicit identification of which claims are contentious: the weight structure implements the suspension automatically. The direction of eventual resolution determines whether the claim was a genuine site of innovation or merely a persistent point of disagreement, and this distinction becomes visible only through continuous tracking of the posterior weight $w^+(\gamma, \sigma)$ over time.
\end{remark}

\section{Trust in the World of AI}

\subsection{AI Agents as Sources}

AI agents are, in the precise sense of this framework, capable but error-prone sources. They possess both the generative ability to produce perceptions $\Gamma_\sigma(\gamma)$ of claims across broad and diverse realms $\mathcal{N}$ and the discriminative capability to form truth assessments $\Theta_\sigma(\Gamma_\sigma(\gamma))$ of those perceptions. Their error-proneness is structural: stochasticity in the generative process, sensitivity to context, and dependence on inaccessible inputs mean that $\Gamma_\sigma(\gamma)$ is neither perfectly reproducible nor fully transparent to external observers. AI agents are therefore neither confirmation-reliable nor straightforwardly verifiable --- they are precisely the class of source for which reputation of conviction, as formalized in this article, is the operative mechanism of trust.

\subsection{Training and Pre-Deployment Certification}

The training process operates largely in the assimilative regime. Objective truths are well established across the domains on which agents are trained, prior certitudes $w^-(\gamma)$ are high, and loss functions measure correctness --- the alignment of the agent's assessment with the established consensus. This is appropriate: in the assimilative regime, correctness is a reliable proxy for conviction, and optimizing for it is both tractable and well-motivated.

Post-training approaches attempt to push agents into the augmentative regime --- to produce perceptions that are not merely correct but genuinely conviction-worthy. The most prominent of these, direct preference optimization, works by contrasting pairwise perceptions and driving the source toward a preferred set. This is directionally aligned with the desiderata of this framework: it orients the agent toward producing perceptions that reviewers gravitate to, which is precisely the human expectation articulated in Section~\ref{sec:reputation}. The limitation is that preference is a subjective and potentially inconsistent proxy for conviction, and the resulting agent may be persuasive without being demonstrable.

Benchmarks have traditionally served as the pre-deployment measure of agent capability. Their value is eroding: sources trained on sufficiently broad corpora have, in effect, encountered benchmark claims during training, making benchmark performance a measure of memorization as much as generalization. More fundamentally, benchmarks assess performance over a fixed realm $\mathcal{R}_{\text{bench}} \subseteq \mathcal{N}$, and the conflation of strong performance in one realm with reliable performance across others is precisely the pointwise failure mode that can become catastrophic in practice. A source may perform well in expectation over $\mathcal{N}$ while being systematically unreliable on the specific claims that matter.

Independent pre-deployment certification against well-defined realms of claims is therefore crucial. Such certification is the analogue of human graduation --- a structured, third-party assessment of a source's conviction over a defined scope, conducted before the source is trusted to operate consequentially. It establishes a baseline reputation $R_\sigma(\mathcal{R})$ prior to deployment and gives consumers a principled basis for the initial extension of trust.

\subsection{Inference and Continuous Reputation}

Pre-deployment certification, however, is not sufficient. A deployed agent is subject to a far broader and less predictable distribution of claims than any training or certification regime can anticipate. Guardrails --- rules, filters, and refusal mechanisms --- are a necessary response to this, but they are inherently reactive and perpetually incomplete. Language alone is extraordinarily expressive --- even in LLM-specific deployments, the space of undesirable situations is effectively infinite, and any finite set of guardrails will be outpaced by the diversity of deployment contexts. The impossibility of zero errors is not a failure of engineering ambition; it is a structural property of deploying capable sources into open-ended realms.

The appropriate response is not to pursue the fiction of a fully guarded agent, but to acknowledge the inevitability of error and build the infrastructure to detect, characterize, and account for it. This is precisely what the reputation framework provides. A reliable AI agent must produce perceptions that are self-sufficient and assessable, so that conviction can be measured by external verifiers as claims arise. It must accumulate a record of such assessments across the full distribution of deployed claims, so that its reputation ${R}_\sigma(\mathcal{R})$ reflects a stable, claim-sensitive, and continuously updated history of vindication --- not a point-in-time declaration. And it must participate in an ecosystem that credits this accumulation and rewards agents whose conviction has proven durable, making the incentive to produce complete and transparent perceptions structural rather than incidental.

This points toward a model of continuous verification, where independent verifiers assess an agent's perceptions against emerging consensus and commit those assessments to a trustless trail. As assessments accumulate, the reputation estimate stabilizes, and trust in the agent becomes estimable rather than assumed. Certification is where trust begins. Reputation is where it is earned. Trust in AI agents is not a property to be declared at a point in time, but one that accrues, is observable, and can be lost.

\section{Conclusion}

This article has developed a mathematical framework for knowledge, truth, and trust grounded in the structure of claims and sources. Truth is defined as the reproducibly perceived subset of knowledge, arising as an asymptotic consensus across independent perceptions. Sources are characterized by their generative and discriminative roles, and their interactions with claims are organized into six bilateral relationships --- faithfulness, conviction, transparency, correctness, neutrality, and redundancy --- of which the first three are unconditional desiderata and the latter three characterize the assimilative regime.

Among these, conviction emerges as the principled basis for reputation. It is regime-independent, penalizes neither innovation nor well-founded dissent, and demands the transparent and self-sufficient perceptions that make external verification tractable. Reputation is formalized as the expected weighted signed conviction over a realm of claims, with claim weights determined by prior and posterior certitude of objectivity. The resulting measure is bounded, claim-sensitive, and continuous --- accruing gradually through a verifiable record rather than through singular declarations.

The application to AI agents closes the argument. Agents are capable but error-prone sources, and the framework identifies continuous verification as the mechanism through which their reputation can be established and maintained. Point-in-time certification is insufficient; what is required is a trustless trail of assessments that accumulates over deployment, makes reputation observable, and allows it to be lost.

\begin{note}
The framework developed here ultimately issues a dual charge. To builders of AI agents: architect for verifiable conviction --- produce systems whose perceptions are complete, whose reasoning is self-sufficient, and whose outputs can be stood behind when challenged. To consumers of AI agents: demand it --- resist the temptation of capable but unverifiable systems, and extend trust only where reputation has been earned. The charge runs in the other direction too. To consumers: commit to supporting the infrastructure of continuous verification. To builders: demand it of the platforms on which your agents are deployed and gain reputation based on merit. Millennia of evolutionary and social pressure have furnished humans with precisely this dual machinery --- the instinct to build reliable reputations and the instinct to consult them. That we would deploy agents of ever-increasing capability into the world without reconstructing it artificially is the central oversight of our moment.
\end{note}

\renewcommand{\refname}{}

\end{document}